\documentclass{eurodyn2026blank}
\usepackage{amsmath}
\usepackage{array}
\usepackage{subfig}
\usepackage{float}

\title{GREEN PHYSICS-INFORMED MACHINE LEARNING MODELS FOR STRUCTURAL HEALTH MONITORING}

\author{Daisy R. Bradley$^1$, Elizabeth J. Cross$^2$}

\heading{Daisy R. Bradley, Elizabeth J. Cross}

\address{$^1$University of Sheffield\\
  Sheffield, UK\\
  e-mail: dbradley2@sheffield.ac.uk \and
  $^2$
  University of Sheffield\\
  Sheffield, UK\\
e-mail: e.j.cross@sheffield.ac.uk}


\abstract{Machine learning continues to emerge as an important tool to be utilised within structural engineering and structural health monitoring, due to its ability to accurately and quickly perform both regression and classification tasks.
However, a purely data driven approach has its limitations, particularly where we lack data from relevant environmental and operational conditions, a situation that has led to the development of physics-informed machine learners for structural health monitoring.
These ``grey-box'' models take into account the physical insight that an engineer would have about the structure they are modelling and have shown promising results in the structural engineering field among many others.
This work compares black and grey-box models through a ``green'' lens, comparing them in terms of their environmental impact, and investigating how the high extrapolative performance of grey-box models can reduce their runtimes and therefore carbon emissions.
The authors aim to develop physics-informed models with reduced computational costs, while maintaining high performance, illustrated through a structural health monitoring case study.}

\begin{document}

\section{INTRODUCTION}
It is inevitable that within the engineering community, our use of machine learning (ML) will significantly increase in the coming years.
As new models are created, likely to be larger and more powerful than the current state of the art \cite{chen_survey_2022}, the energy demand of engineering ML will increase accordingly, alongside its environmental impact.
The computer science research community have published several works on reducing the emissions associated with ML, as discussed below in the literature review.
However, within the engineering sector, the majority of research on sustainable ML focuses on implementing ML as a tool to achieve sustainability related goals, rather than the sustainability of models themselves.
This work aims to start to address this gap in the literature by exploring the sustainability of ML models within an engineering (structural health monitoring (SHM)) context. 
In particular, we begin to explore the question of how the environmental impact of a model changes with the addition of physical knowledge, by comparing a purely data driven ``black-box'' against two physics-informed machine learning (PIML) or ``grey-box'' models with different levels of integrated physical knowledge. The authors explore the working hypothesis that reduced emissions are possible when the physical knowledge embedded in the model allows a reduction in training data without significantly increasing model complexity.

\section{LITERATURE REVIEW}

Research on the emissions produced by ML models has identified several contributing factors, such as the choice of hardware, the use and power usage effectiveness (PUE) of a data centre, the carbon intensity of the electricity generated to run the model/data centre, and the time taken for the model to run/train \cite{paul_comprehensive_2023, bolon-canedo_review_2024, gupta_chasing_2020, luccioni_estimating_nodate, lannelongue_green_2021, luccioni_counting_2023}.
Within the field of sustainable ML, it is well understood that having an awareness of our emissions is the first step to reducing them \cite{lannelongue_ten_2021} and as a result, several packages and programmes have been created to quantify carbon emissions from computing, including CodeCarbon \cite{benoit_courty_2024_11171501}, which will be used throughout this work to compare models using an estimation of their environmental footprints.
There is extensive work in the literature on increasing the efficiency of ML models, though this is rarely solely motivated by environmental factors.
In Menghani's 2023 review, the author categorises efficient techniques into five focus areas: compression techniques, learning techniques, automation, efficient architectures, and infrastructure and hardware \cite{Menghani_review_2023}. 
In large language models (LLMs), some of the most successful techniques include pruning, quantisation, and mixture of expert (MoE) architectures.

Many traditional engineering models rely on numerical approaches, where there are a number of means of increasing efficiency (e.g. reduced order and surrogate models). 
Our focus here is on reducing the burden of ML models employed for engineering tasks where less attention has been given.
We look particularly to the field of PIML as a promising field to provide methods with reduced emissions.
Physics-informed machine learners are a subcategory of ML models, where physical knowledge is embedded into the black box to create a hybrid grey box. 
Examples include residual models, hybrid architectures, constrained machine learners, and the popular physics-informed neural networks (PINNs) \cite{Cross_2021, Karniadakis2021PhysicsInformedML}.
Some of the benefits of grey-box models compared to their black-box counterparts include increased generalisation, efficiency and interpretability \cite{Meng2025WhenPhysicsMeetsML}. 
Although PIML models have received much attention in the literature, this is the first work to introduce physics knowledge into a model with the intention of reducing its environmental footprint.
As an initial study on the topic, we develop and demonstrate a very simple set of test cases, using a Gaussian Process (GP) model structure that can be easily interpreted and interrogated.
We choose GPs for this initial study not because of their heavy computational burden (the examples here are in fact particularly light), but because they provide a framework where physics can be embedded in multiple different ways without changing the model architecture \cite{Cross2024GPs}, providing a direct means of comparing emissions with varying levels of embedded physics.

\section{TEST CASES}

It is well known that grey-box models often require less training data than black-box models, especially in extrapolative scenarios. 
This is demonstrated in Figure 1 for a generated surface, where a black-box model (a standard GP regression) is unable to perform adequately outside of its training set, whereas the grey-box model (with insight embedded in a kernel) achieves a far better performance with the same training data. 

\begin{figure}[H]
\centering

\subfloat[Black-1 - NMSE 72.1]{
    \includegraphics[width=0.32\textwidth]{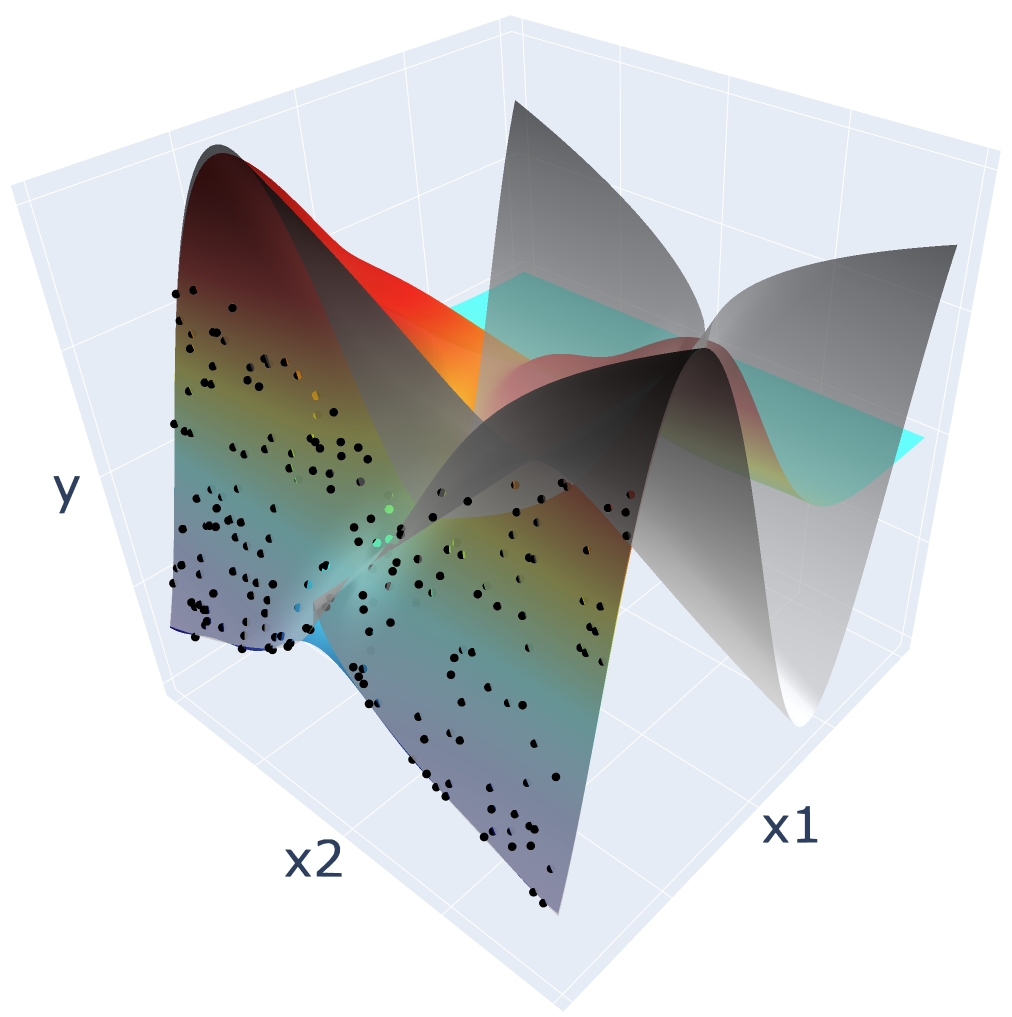}
}
\hspace{0.02\textwidth}
\subfloat{
  \raisebox{0.5cm}{
    \includegraphics[width=0.15\textwidth]{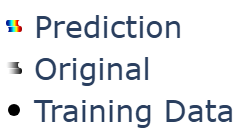}
  }
}
\hspace{0.02\textwidth}
\subfloat[Grey-2 - NMSE 2.92]{
    \includegraphics[width=0.32\textwidth]{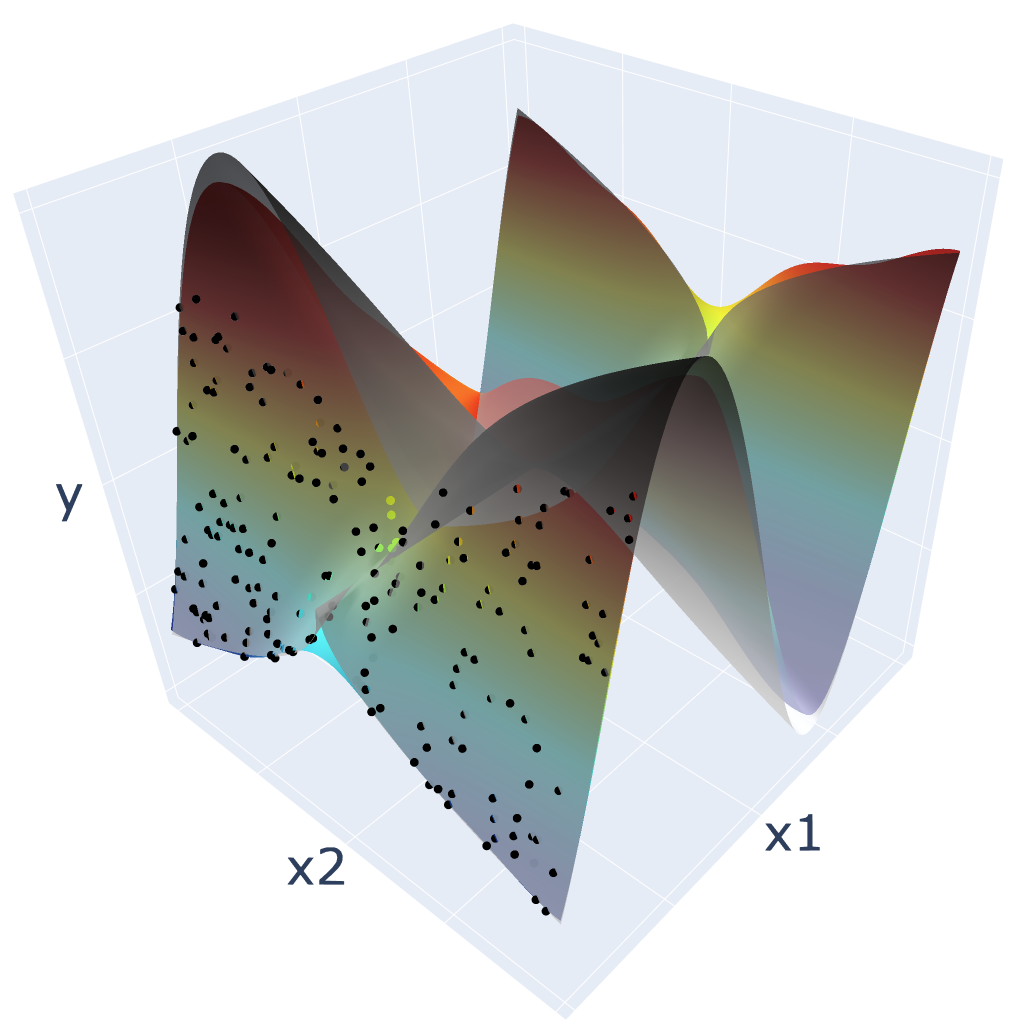}
}

\caption{The performance of black-box and grey-box models with 20\% training data coverage, where NMSE is a normalised mean squared error}
\end{figure}

Although the requirement for training data often decreases when embedding physics into a model, model complexity often increases in return, creating a complex relationship between the amount of physics and the training time (hence energy consumption and emissions) of a model. The surface seen in Figure 1, and defined by Equation 1, is a very simple ``toy-box" test case, which was used as a starting point to explore this trade-off.

\begin{eqnarray}
y = \sqrt{|x_2|}\,\sin\!\left(4\,x_1\right)
\end{eqnarray}

To explore emissions, three GP models were designed with various ``greyness", where the embedded information/knowledge concerns the periodicity of the surface. 
The first model, Black-1, is a commonly used black-box GP regression, with a constant mean function and a squared exponential (SE) covariance function (kernel), which is defined in Equation 2. 
This kernel consists of two hyperparameters, length-scale ($l$) and output variance ($\sigma^2$), hence Black-1 has a total of four hyperparameters to learn - two from the kernel, the mean, and the observation noise. For all models, learning is achieved by maximising the marginal likelihood \cite{RasmussenWilliams2006}.

\begin{eqnarray}
k_{SE}(x,x') = \sigma^2exp(-\frac{(x-x')^2}{2l^2})
\end{eqnarray}

The two grey-box models also use a constant mean function, but differ to Black-1 by multiplying the SE kernel by a periodic kernel,  outlined in Equation 3, which we use as our physics knowledge proxy. 

\begin{eqnarray}
k_{Per}(x,x') = \sigma^2exp(-\frac{2sin^2(\pi |x-x'|/p)}{l^2})
\end{eqnarray}

The periodic kernel consists of 3 hyperparameters - lengthscale ($l$), output variance ($\sigma^2$), and period ($p$). 
The grey-box models have a total of 6 hyperparameters - the lengthscale and period from the periodic kernel, the lengthscale from the SE kernel, an overall output variance, the mean, and the observation noise.
To mimic scenarios of more or less information available on system parameters, in the first grey-box model, Grey-1, the period is bounded within 10\% of the true value, and in the second grey-box model, Grey-2, the period is fixed at the true value and therefore not optimised in the training process.
Therefore, Grey-2 is ``whiter'' than Grey-1, and has one less hyperparameter to learn. 
A description of the three models is shown in Table 1.

\begin{table}[h]
  \begin{center}
    \begin{tabular}{{cccc}}
      \toprule
       & Black-1 & Grey-1 & Grey-2\\
      \midrule
      Mean function     & Constant  & Constant & Constant    \\
      Covariance function    & SE     & SE*Periodic & SE*Periodic  \\
      Number of hyperparameters to be optimised     & 4     & 6 & 5     \\
      \bottomrule
    \end{tabular}
  \end{center}
  \caption{Model details for Black-1, Grey-1, and Grey-2}
\end{table}

One of the advantages we have in our mission to reduce emissions in an engineering context is that we do, in many cases, have some idea of the level of error required for a particular modelling task.
We are able to say, perhaps more than in other disciplines, how good ``good enough” actually is. 
This means that we can set targets on errors required to limit our computing effort - rather than just design algorithms to be efficient, we can be purposefully ``green” by explicitly considering the trade-off  between model error and computational burden. 
Therefore, in this work, the models were compared based on the training data coverage that was required to give a normalised mean squared error (NMSE) below 10. 
Coverage was varied in 10\% intervals across the $x_1$ variable, with 100 training data points per 10\%.
The final models used in forthcoming carbon calculations corresponded to the lowest coverage for which no NMSE values exceeded 10 across nine runs for hyperparameter optimisation (three repeats for three start points) at each interval.

The models were created using the GPyTorch package (V1.14.2) in Python \cite{gardner2018gpytorch}.
To train the models, the ADAM gradient descent optimiser \cite{kingma2017adammethodstochasticoptimization} was used with 1000 iterations.  
Codecarbon \cite{benoit_courty_2024_11171501} was used to estimate the carbon emissions from each run, which were conducted on a Dell Inspiron 5680, with 32.0 GB of random-access memory (RAM) and an Intel Core i7-8700 central processing unit (CPU) with twelve cores in total. 
Note that CodeCarbon \cite{benoit_courty_2024_11171501} was unable to access live data from the components, so the graphics card was not used, and the emissions data was estimated based on the thermal design power (TDP) of the CPU, as outlined in the CodeCarbon methodology \cite{codecarbon_methodology}.
While the footprints of these models will be very small, this work aims to form the foundation of a concept that can be scaled up to much larger models, where carbon savings become much more significant.

\subsection{An engineering test case}

The same methodology described above was applied to a set of engineering data, collected for use in SHM.
SHM is a condition-based maintenance scheme which aims to increase the safety of structures and prevent the early retirement of healthy components.  
SHM uses ML to predict the health state of a structure based on its dynamic response, however, the response of a damaged structure may be similar to that of a healthy structure under particular environmental or operational conditions (EOCs). 
It is commonplace that data cannot be collected across all EOCs, especially when measurements can only be made in a certain time window, requiring any algorithm in use to be able to function in unseen conditions.

The SHM dataset used here is from a lab-based modal test of a simple metallic aircraft structure - the GARTEUR benchmark, shown in Figure 2 (see \cite{Link2003} and \cite{Gibson2024PIML} for specific details). 
For this initial work, the wing, instrumented with accelerometers, was excited via an electrodynamic shaker to produce a response in its first mode (close to 10Hz), producing a similar learning surface to that of the toy-box example above - see Figure 3.

\begin{figure}[H]
\centering
 \includegraphics[width=0.7\textwidth]{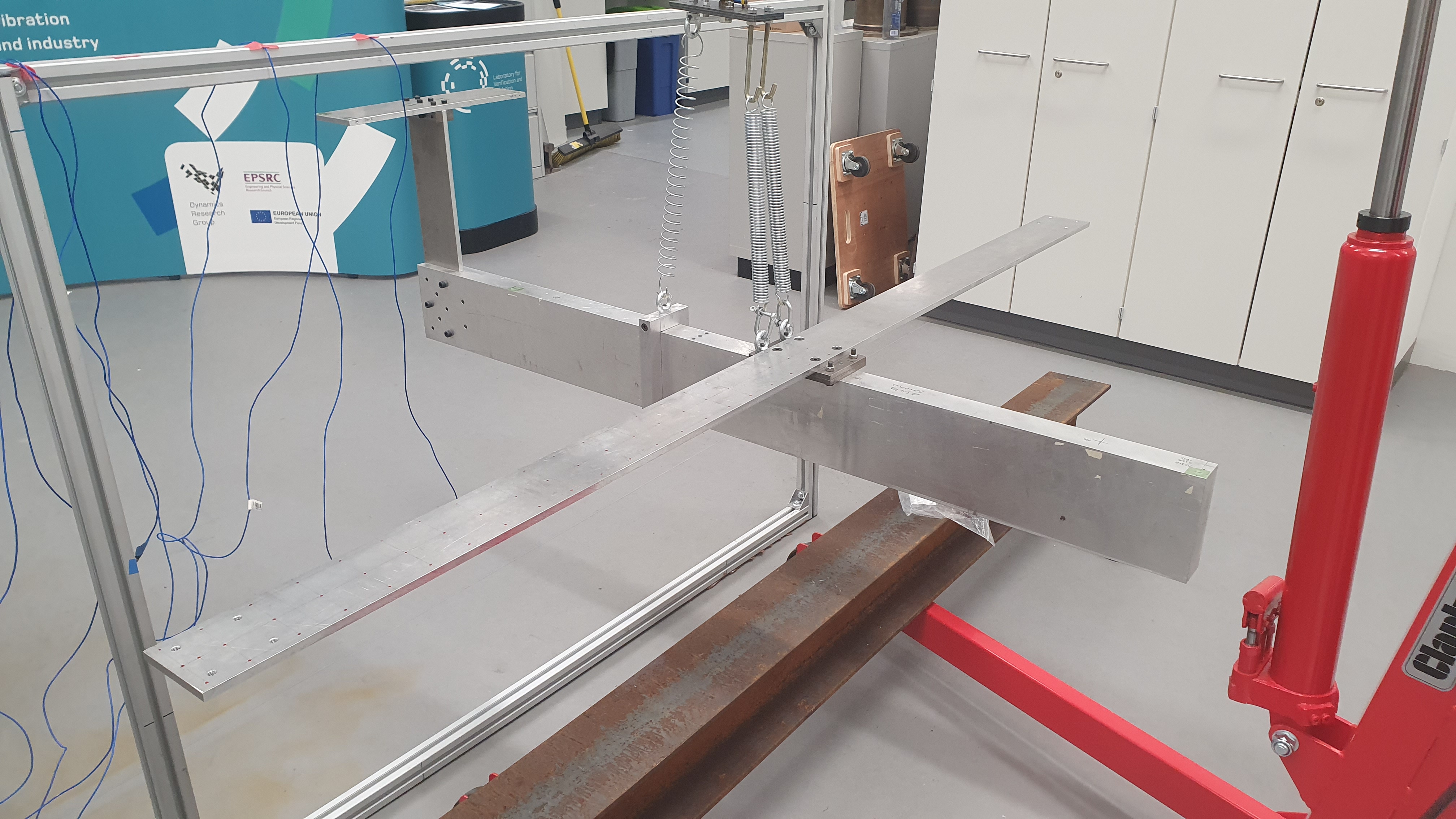}
\caption{The GARTEUR structure used to collect the engineering test case data}
\end{figure}

The ML task for this initial study is to predict the acceleration response spatially and temporally, with the same periodic kernel employed as above used to bring in ``knowledge" of the periodic nature of the modal response. 
Two test cases were employed: the first using the accelerometer positions and every tenth data point, the second using an upsampled dataset, as can be seen in Figure 3, such that each 10\% of surface coverage contains 200 training data points.

\begin{figure}[H]
\centering

\subfloat[Original]{
    \includegraphics[width=0.4\textwidth]{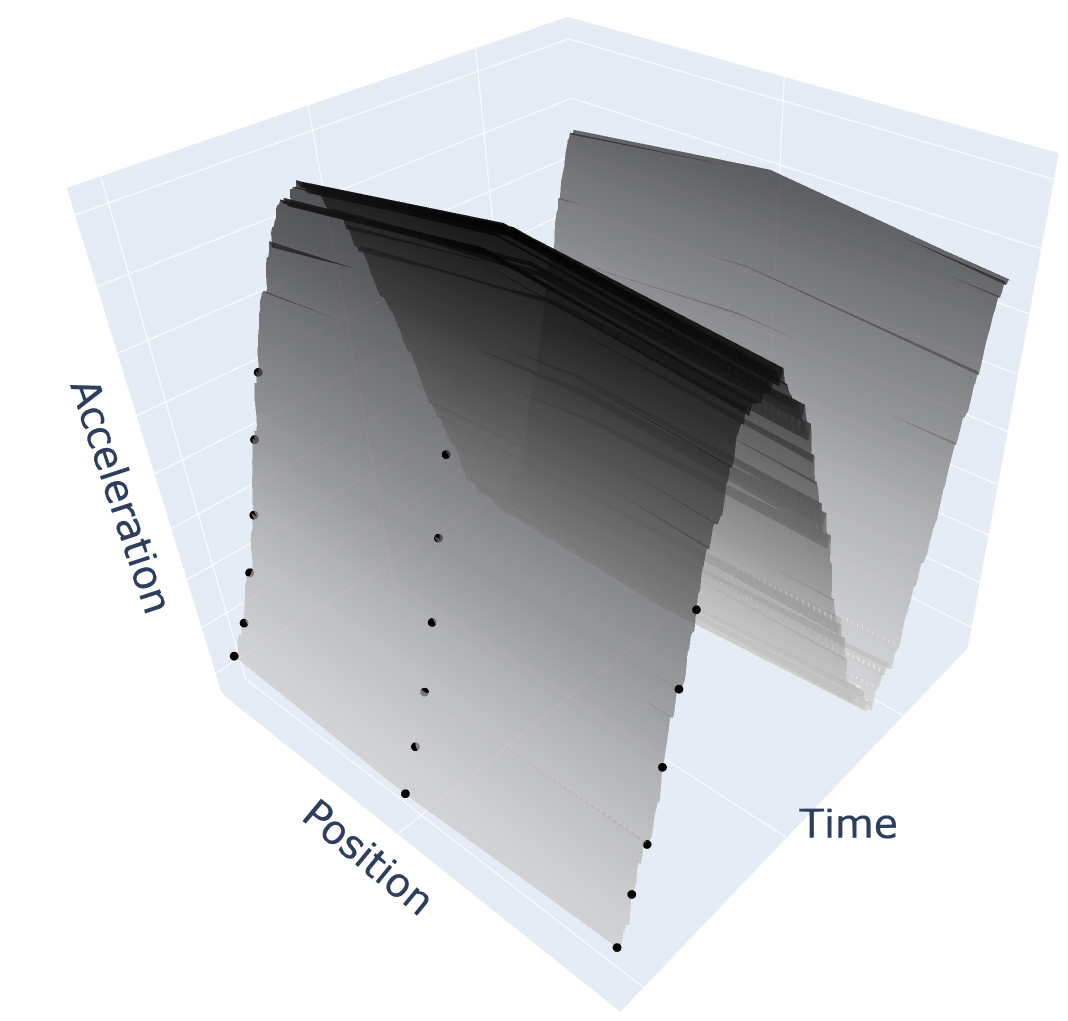}}
\hspace{0.02\textwidth}
\subfloat[Upsampled]{
    \includegraphics[width=0.4\textwidth]{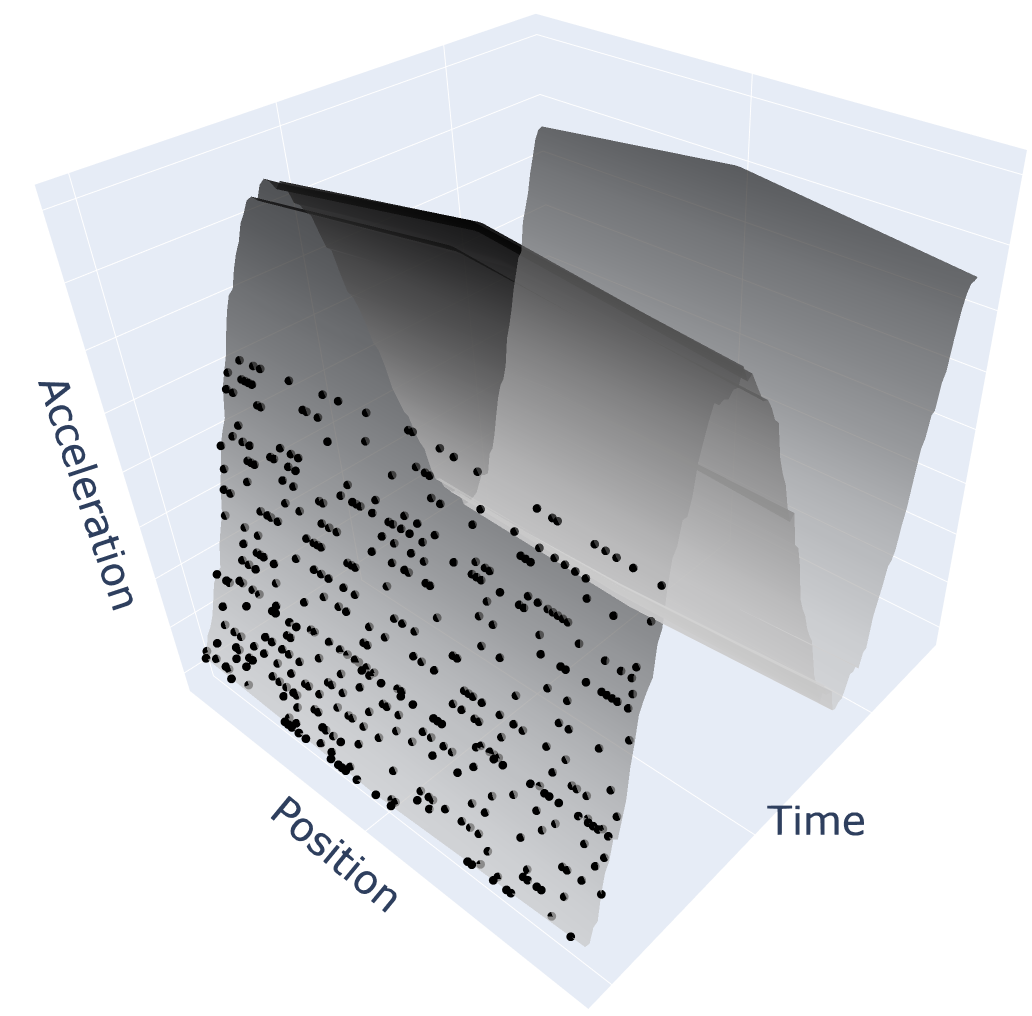}}

\caption{The different training data layouts used for the GARTEUR test cases with 20\% training data coverage}
\end{figure}

\section{RESULTS} 

As expected, the amount of training data required to reach the threshold performance metric decreased as more physics knowledge was added into the models. 
For the toy-box study, the coverage and CodeCarbon \cite{benoit_courty_2024_11171501} results are shown in Table 2. 

\begin{table}[H]
  \begin{center}
    \begin{tabular}{{cccc}}
      \toprule
      & Black-1 & Grey-1 & Grey-2 \\
      \midrule
      Training data coverage (\%)  & 80 & 60 & 20  \\
      $\Delta$ to Black-1 & - & - 25.0\% & - 75.0\% \\
      Emissions ($gCO_{2}e$)  & 0.445 & 0.420 & 0.123 \\
      $\Delta$ to Black-1 & - & - 5.68\% & - 72.4\% \\ 
      \bottomrule
    \end{tabular}
  \end{center}
  \caption{Training data coverage required to meet the performance threshold and CodeCarbon \cite{benoit_courty_2024_11171501} emissions for Black-1, Grey-1 and Grey-2 on the toy-box case study}
\end{table}

A similar trend was seen in the engineering examples, where less coverage was required as more knowledge is embedded.
The coverage and CodeCarbon \cite{benoit_courty_2024_11171501} results can be found in Table 3 for the original layout, and Table 4 for the upsampled layout.

\begin{table}[H]
  \begin{center}
    \begin{tabular}{{cccc}}
      \toprule
      & Black-1 & Grey-1 & Grey-2\\
      \midrule
      Coverage (\%)  & 70 & 60 & 30 \\
      $\Delta$ to Black-1 & - & - 14.3\% & - 57.1\% \\
      Emissions ($gCO_{2}e$)  & 0.0746 & 0.121 & 0.0984 \\
      $\Delta$ to Black-1 & - & + 62.2\% & + 31.9\% \\ 
      \bottomrule
    \end{tabular}
  \end{center}
  \caption{Training data coverage to meet the performance threshold and CodeCarbon \cite{benoit_courty_2024_11171501} emissions for Black-1, Grey-1 and Grey-2 on the GARTEUR data (original layout)}
\end{table}

\begin{table}[H]
  \begin{center}
    \begin{tabular}{{cccc}}
      \toprule
      & Black-1 & Grey-1 & Grey-2\\
      \midrule
      Coverage (\%)  & 80 & 60 & 20 \\
      $\Delta$ to Black-1 & - & - 25.0\% & - 75.0\% \\
      Emissions ($gCO_{2}e$)  & 2.04 & 3.38 & 0.209 \\
      $\Delta$ to Black-1 & - & + 65.8\% & - 89.7\% \\ 
      \bottomrule
    \end{tabular}
  \end{center}
  \caption{Training data coverage to meet the performance threshold and CodeCarbon \cite{benoit_courty_2024_11171501} emissions for Black-1, Grey-1 and Grey-2 on the GARTEUR data (upsampled)}
\end{table}

\section{DISCUSSION}

The emissions calculated here with CodeCarbon \cite{benoit_courty_2024_11171501} used constants for carbon intensity, TDP, and power associated with RAM, meaning changes in emissions are caused by changes in runtime. 
Hence, the discussion will be framed in terms of runtime. 

In a Gaussian process regression we expect that runtime is dominated by a term proportional to the training data size ($N$) cubed, with a secondary term from the gradient descent proportional to the number of hyperparameters to be optimised ($H$) and training data squared \cite{RasmussenWilliams2006}, making the computation $\sim \mathcal{O}(N^3)+\mathcal{O}(HN^2)$. In a standard GP implementation, we therefore expect that when $N$ is large, the impact of the hyperparameter optimisation on runtime is insignificant, though it is more prominent at small $N$, especially in complex models with many hyperparameters. We note again that in this initial exploratory work both hyperparameter counts and training data sizes are small in themselves.

Firstly, let us consider the toy-box example, which was designed as an idealised test case.
Figure 4 shows the impact of training data coverage on runtime for the three models. 
All models show an increasing, approximately quadratic, relationship between runtime and training data coverage, with Grey-1 having the longest runtime at a given coverage, followed by Grey-2 and Black-1 respectively.
This is likely caused by Grey-1 having the most hyperparameters (six), followed by Grey-1 (five), and Black-1 having the fewest (four), and highlights that adding physics to a model can increase runtime when there is no training data reduction. 
However, the toy-box data provides an example where the increase in model complexity is overcome by the runtime reduction from reducing training data, such that the runtime of the physics-informed models is less despite having more hyperparameters.
This data emphasises the importance of reducing training data to decrease model emissions.

\begin{figure}[H]
    \begin{center}
      \includegraphics[width=14cm]{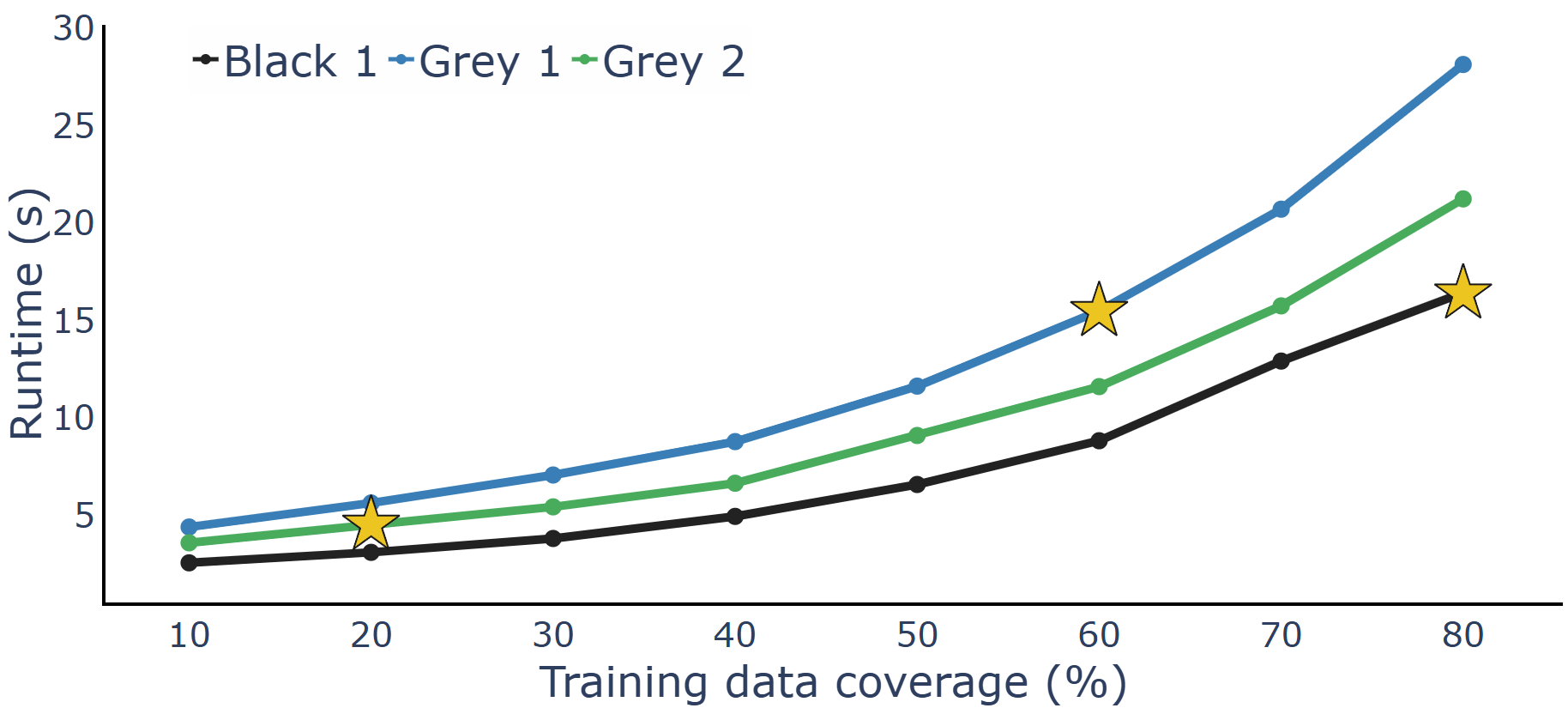}
      \caption{The relationship between training data coverage and runtime for Black-1, Grey-1 and Grey-2 on the toy-box data. The stars indicate the model where the 10\% training target has been met}
    \end{center}
\end{figure}

Figure 5 shows the same study conducted on the upsampled GARTEUR data.
This study has a training data density twice that of the toy-box study, and so it is unsurprising that the first half of this figure is very similar to Figure 4 (as the total amount of data points is the same). 
However, in the second half of Figure 5, we see some larger differences in runtimes between the models, with Grey-1 particularly divergent. 
Once more, this highlights the trade-off between hyperparameter count and training data size, though it is clear that further investigation is needed to quantify this trade-off even for this simple benchmark. 
As expected, the results suggest that runtime (and consequently carbon) savings would increase substantially with larger training data sizes, highlighting the potential benefits when scaling to industry-sized models.

\begin{figure}[H]
    \begin{center}
      \includegraphics[width=14cm]{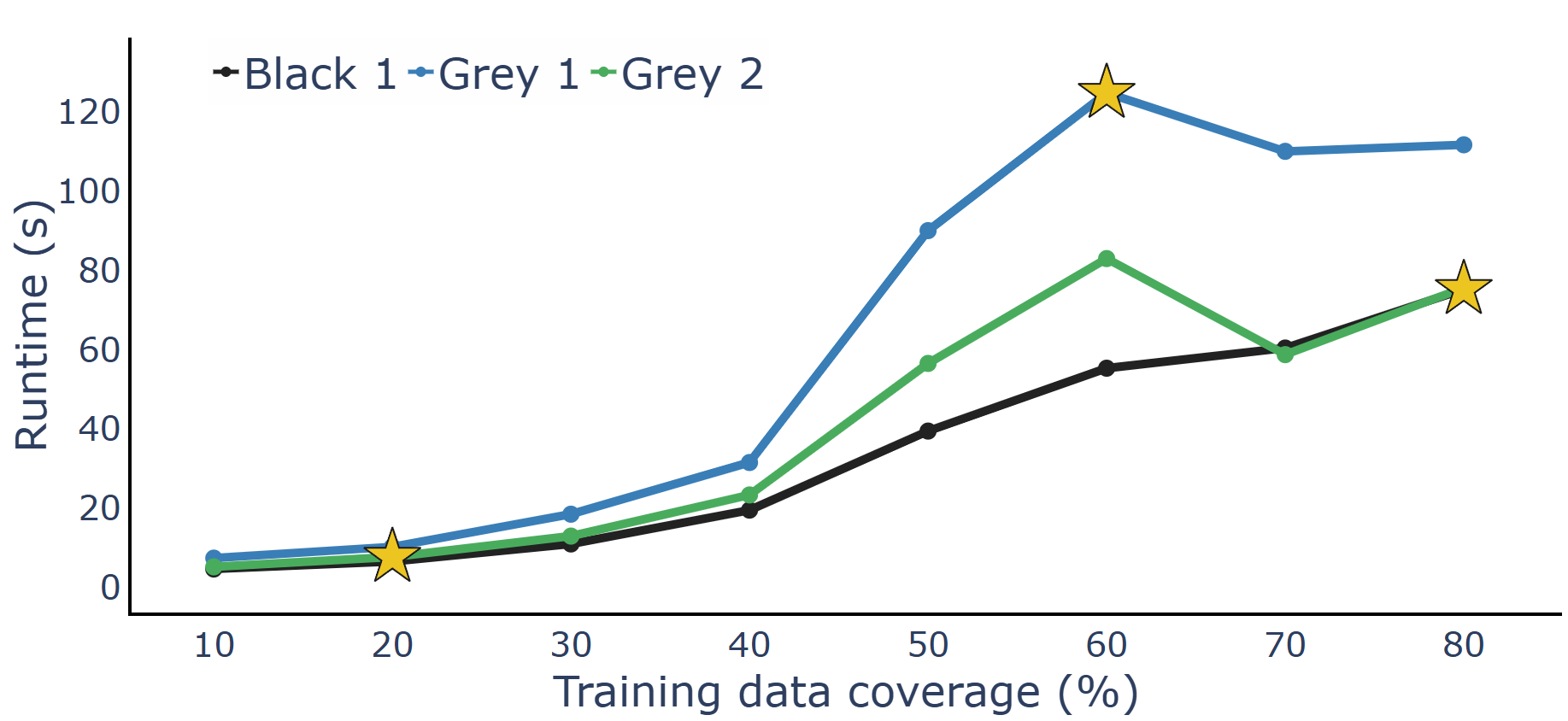}
      \caption{The relationship between training data coverage and runtime for Black-1, Grey-1 and Grey-2 on the upsampled GARTEUR data. The stars indicate the model where the 10\% training target has been met}
    \end{center}
\end{figure}

Finally, we consider the original GARTEUR data, shown in Figure 6.
Here, we see the same pattern as previously for a given coverage, however little trend between coverage and runtime. 
This is due to the very small training data sizes used, with a sparse density within each 10\% coverage.
This data reinforces that when training data reduction is very small (63 points at 70\% coverage vs 27 points at 30\% coverage), the hyperparameter count is more significant in determining runtime.

\begin{figure}[H]
    \begin{center}
      \includegraphics[width=14cm]{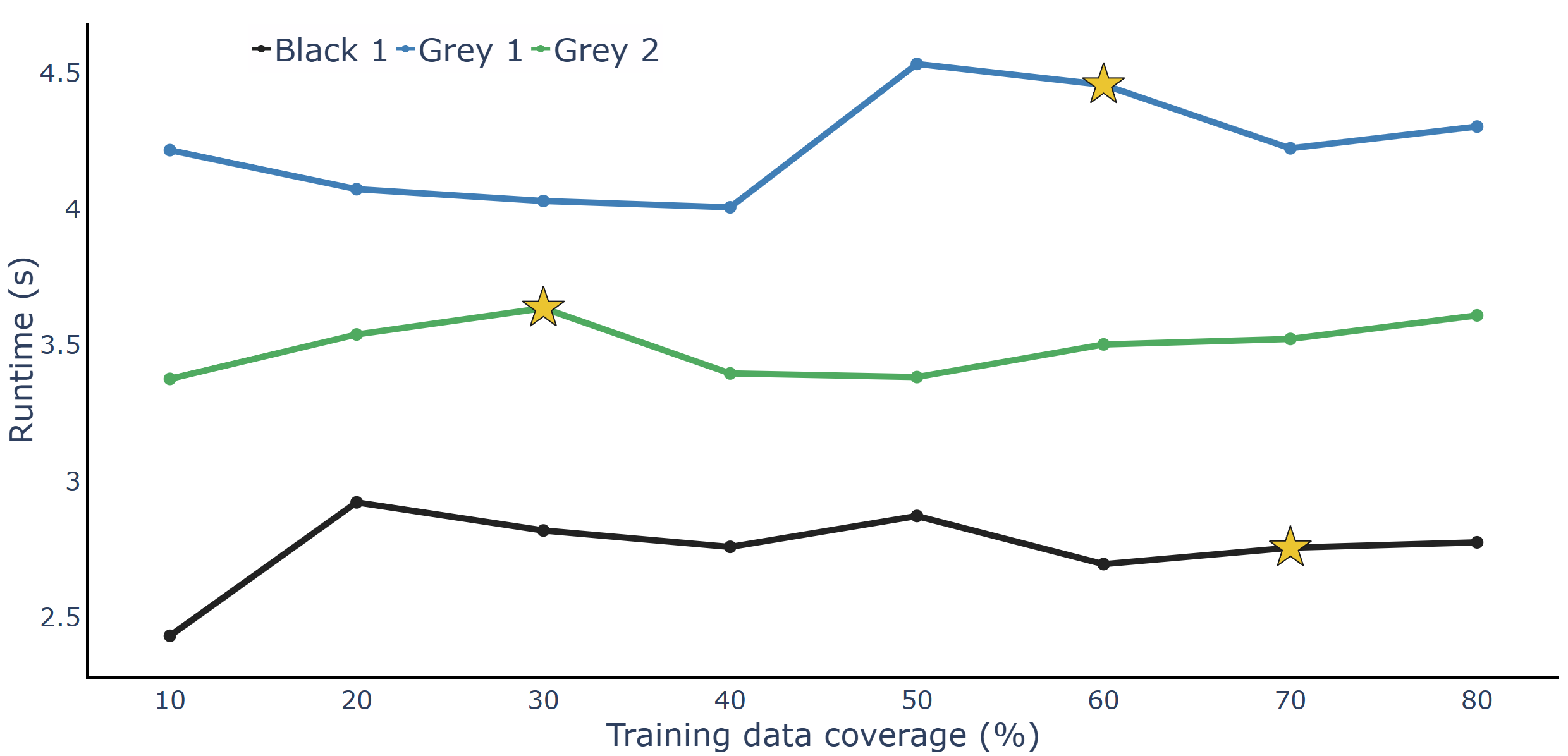}
      \caption{The relationship between training data coverage and runtime for Black-1, Grey-1 and Grey-2 on the original GARTEUR data. The stars indicate the model where the 10\% training target has been met}
    \end{center}
\end{figure}

This work was designed to provide a preliminary investigation into the relationship between PIML, training data and emissions. 
As such, we have only investigated very simple test cases, using simple lightweight GP models with a periodic kernel acting as a proxy for physical insight.
From these results, it appears that the carbon savings become more apparent with large training data sizes, suggesting that the benefits of PIML may be more pronounced in larger, industry-scale models.
While the results thus far are promising, further work must be undertaken before reliable conclusions can be made.
Such work includes upscaling to significantly higher training data sizes and parameter counts, and exploring different models, both in terms of more complex GPs, physics and beyond. 
Neural networks and PINNs are of particular interest, both due to their popularity and linear scaling laws with both hyperparameters and training data.

\section{CONCLUSION}

This work is a preliminary study investigating the impact of introducing physics on model emissions via decreasing training data requirements.
Three GP models were designed with various levels of knowledge embedded and tested on two simple case studies.
Fewer training data points were required to meet the performance threshold as physics input increased, though this did not consistently correspond to carbon savings, due to additional hyperparameters in the physics-informed models.
This initial investigation suggests that in Gaussian process regression, physics knowledge is more likely to reduce emissions when working with large amounts of training data.
Future work aims to quantify this trade-off and explore whether more complex GPs and other, industry scale ML models observe a similar trend.

\bibliographystyle{plainnat}
\bibliography{references}

\end{document}